\documentclass{article}
\usepackage{spconf,amsmath,graphicx}
\usepackage{algorithm,algpseudocode}
\usepackage{mathtools}
\usepackage{cancel}
\usepackage{amssymb}
\usepackage{booktabs}
\usepackage{caption}
\usepackage{subcaption}
\usepackage{multirow}
\usepackage{setspace}
\usepackage{icomma}

\usepackage{pifont}
\usepackage{tikz}
\usetikzlibrary{positioning}
\usetikzlibrary{calc}
\usepackage{pgfplots, pgfplotstable}    
\pgfplotsset{compat=1.17} 
\usepackage{url}
\DeclareMathSymbol{\shortminus}{\mathbin}{AMSa}{"39}

\DeclareMathOperator*{\argmin}{argmin}
\newcommand{\Sref}[1]{\S\ref{#1}}
\makeatletter
\def\blfootnote{\xdef\@thefnmark{}\@footnotetext}
\makeatother

\usepackage[
backend=biber,
style=ieee,
citestyle=numeric-comp,
maxbibnames=3,
maxcitenames=3,
doi=false,isbn=false,url=false,eprint=false
]{biblatex}

\defbibheading{bibliography}[\refname]{}

\DeclareSourcemap{
	\maps[datatype=bibtex, overwrite=true]{
		\map{
			\step[fieldsource=booktitle,
			match=\regexp{.*Interspeech.*},
			replace={Proc. Interspeech}]
			\step[fieldsource=journal,
			match=\regexp{.*INTERSPEECH.*},
			replace={Proc. Interspeech}]
			\step[fieldsource=booktitle,
			match=\regexp{.*ICASSP.*},
			replace={Proc. ICASSP}]
			\step[fieldsource=booktitle,
			match=\regexp{.*icassp_inpress.*},
			replace={Proc. ICASSP (in press)}]
			\step[fieldsource=booktitle,
			match=\regexp{.*Acoustics,.*Speech.*and.*Signal.*Processing.*},
			replace={Proc. ICASSP}]
			\step[fieldsource=booktitle,
			match=\regexp{.*International.*Conference.*on.*Learning.*Representations.*},
			replace={Proc. ICLR}]
			\step[fieldsource=booktitle,
			match=\regexp{.*International.*Conference.*on.*Computational.*Linguistics.*},
			replace={Proc. COLING}]
			\step[fieldsource=booktitle,
			match=\regexp{.*SIGdial.*Meeting.*on.*Discourse.*and.*Dialogue.*},
			replace={Proc. SIGDIAL}]
			\step[fieldsource=booktitle,
			match=\regexp{.*International.*Conference.*on.*Machine.*Learning.*},
			replace={Proc. ICML}]
			\step[fieldsource=booktitle,
			match=\regexp{.*North.*American.*Chapter.*of.*the.*Association.*for.*Computational.*Linguistics:.*Human.*Language.*Technologies.*},
			replace={Proc. NAACL}]
			\step[fieldsource=booktitle,
			match=\regexp{.*Empirical.*Methods.*in.*Natural.*Language.*Processing.*},
			replace={Proc. EMNLP}]
			\step[fieldsource=booktitle,
			match=\regexp{.*Association.*for.*Computational.*Linguistics.*},
			replace={Proc. ACL}]
			\step[fieldsource=booktitle,
			match=\regexp{.*Automatic.*Speech.*Recognition.*and.*Understanding.*},
			replace={Proc. ASRU}]
			\step[fieldsource=booktitle,
			match=\regexp{.*Spoken.*Language.*Technology.*},
			replace={Proc. SLT}]
			\step[fieldsource=booktitle,
			match=\regexp{.*Speech.*Synthesis.*Workshop.*},
			replace={Proc. SSW}]
			\step[fieldsource=booktitle,
			match=\regexp{.*workshop.*on.*speech.*synthesis.*},
			replace={Proc. SSW}]
			\step[fieldsource=booktitle,
			match=\regexp{.*Advances.*in.*neural.*information.*processing.*},
			replace={Proc. NeurIPS}]
			\step[fieldsource=booktitle,
			match=\regexp{.*Advances.*in.*Neural.*Information.*Processing.*},
			replace={Proc. NeurIPS}]
			\step[fieldsource=booktitle,
			match=\regexp{.*Workshop.*on.* Applications.* of.* Signal.*Processing.*to.*Audio.*and.*Acoustics.*},
			replace={Proc. WASPAA}]
			\step[fieldsource=publisher,
			match=\regexp{.+},
			replace={{}}]
			\step[fieldsource=month,
			match=\regexp{.+},
			replace={{}}]
			\step[fieldsource=location,
			match=\regexp{.+},
			replace={{}}]
			\step[fieldsource=address,
			match=\regexp{.+},
			replace={{}}]
			\step[fieldsource=organization,
			match=\regexp{.+},
			replace={{}}]
		}
	}
}

\title{Joint Modelling of Spoken Language Understanding Tasks with Integrated Dialog History}
\name{Siddhant Arora${}^1$, Hayato Futami${}^2$, Emiru Tsunoo${}^2$, Brian Yan${}^1$, Shinji Watanabe${}^1$}
\address{${}^1$Carnegie Mellon University, ${}^2$Sony Group Corporation, Japan}
\begin{document}
\ninept
\maketitle
\begin{abstract}
Most human interactions occur in the form of spoken conversations where the semantic meaning of a given utterance depends on the context. Each utterance in spoken conversation can be represented by many semantic and speaker attributes, and there has been an interest in building Spoken Language Understanding (SLU) systems for automatically predicting these attributes. Recent work has shown that incorporating dialogue history can help advance SLU performance. However, separate models are used for each SLU task, leading to an increase in inference time and computation cost. Motivated by this, we aim to ask: can we jointly model all the SLU tasks while incorporating context to facilitate low-latency and lightweight inference? To answer this, we propose a novel model architecture that learns dialog context to jointly predict the intent, dialog act, speaker role, and emotion for the spoken utterance. 
Note that our joint prediction is based on an autoregressive model and we need to decide the prediction order of dialog attributes, which is not trivial. To mitigate the issue, we also propose an order agnostic training method.
Our experiments show that our joint model achieves similar results to task-specific classifiers and can effectively integrate dialog context to further improve the SLU performance.\footnote{Our code \& models are publicly available as part of ESPnet-SLU toolkit.}
\end{abstract}
\begin{keywords}
spoken language understanding, spoken dialog system, end-to-end systems, joint modelling, speaker attributes
\end{keywords}
\section{Introduction}
\label{sec:intro}
Spoken dialogue systems aim to enable dialogue agents to engage in a more natural conversation with humans. They have commonly represented a possible dialogue by a series of frames~\cite{de2008spoken,allen2001toward}. Each frame represents the type of task the user seeks and has attributes representing the information that can help the system to complete the task. These spoken dialog systems aim to automatically identify the topic of conversations as well as other dialog frame attributes (e.g., dialogue act, emotion) to engage in conversation with the user.

Conventional Spoken Language Understanding (SLU)~\cite{agrawal2020tie,cha2021speak} systems independently process each utterance in a conversation. However, the meaning of an utterance in a spoken dialog depends on the context. Prior work has shown dialogue context to be particularly useful in resolving ambiguities and co-references~\cite{bhargava2013easy,xu2014contextual}. As a result, there has been extensive work on incorporating context to improve Natural Language Understanding (NLU) performance~\cite{colombo2020guiding,bothe2018context,raheja2019dialogue}. Prior work~\cite{kim-etal-2019-gated,kim19cross,hori2020transformer} on conversational speech has also shown strong improvements in ASR performance by incorporating dialog context. Thus, several approaches~\cite{chen2016end,sankar2019neural,bapna2017sequential,vukotic2016step} have looked into incorporating dialog history in pipeline based SLU systems. Recently, there has been some work~\cite{ganhotra2021integrating,sunder2022towards,tomashenko2020dialogue} in incorporating context to improve end-to-end (E2E) SLU performance. One such approach~\cite{ganhotra2021integrating} integrates dialog history into an E2E SLU system by using ASR transcripts of previous spoken utterances. There has also been an effort~\cite{sunder2022towards} to incorporate context directly from the audio of previous utterances.

However, these works mostly focus on building a single model for each SLU task like intent classification, dialog act classification, and emotion recognition. To jointly predict all the SLU tasks, we have to execute all models separately. Consequently, these models not only
have a large memory footprint but also have high latency, which can affect the naturalness of spoken conversations~\cite{arora2022two} when these systems are deployed in commercial applications like voice assistants. Prior work has shown that jointly modeling different speech processing tasks together in a united framework can perform comparable to task-specific models~\cite{o2021conformer,zhang2021end} while reducing latency~\cite{chang2019joint}. Motivated by this work, we ask the following questions: 
(i) Can we jointly model all SLU tasks while incorporating context in a single unified implementation without much loss in performance? (ii) Does incorporating the SLU tags predicted at previous spoken utterances by the joint model help in better modeling spoken dialogue context? We seek to answer these questions by proposing a novel E2E SLU architecture that jointly models all the SLU tasks while effectively learning context from previous spoken utterances.
This joint model uses an autoregressive decoder to predict dialog attributes one by one and the order in which the model predicts the SLU attributes may impact model performance. Hence, we investigate (iii) if we can use order agnostic training to make the model automatically predict in the optimal ordering during inference. Inspired by prior work on ASR~\cite{lee2021intermediate}, we also investigate (iv) if an intermediate CTC loss advances SLU performance further.
We conduct extensive experiments on the recently released HarperValley~\cite{wu2020harpervalleybank} Bank dataset, which consists of dialogs between users and consumer bank agents. Our results show that our joint model performs at par with the individual classifiers for each SLU task, and incorporating dialog context and order agnostic training can further lead to significant improvements in performance.
Our code and models are made publicly available as part of the ESPnet-SLU~\cite{ESPnet-SLU} toolkit.

The key contributions of our work are summarised below.
\begin{itemize}
    \item We propose a novel joint model that uses dialog context to jointly predict the intent, dialog act, speaker role and emotion.
    \item We propose order agnostic training of the joint model and show an improvement in performance across all SLU tasks.
    \item We investigate the efficacy of using SLU tags predicted from previous utterances to model dialog context.
    \item We show that incorporating the usage of intermediate CTC loss can advance SLU performance. 
\end{itemize}

\section{Background: Dialog Context in SLU}

\label{sec:slu}
The formulation for SLU with integrated dialog context extends the well-studied framework of NLU systems~\cite{colombo2020guiding,bothe2018context,raheja2019dialogue}. 
For NLU systems, dialog context sequence is represented as sequence of $C$ utterances, i.e., $S = \{s_c| c=1,\dots , C\}$. Each utterance in the dialog context sequence is represented as $s_c = \{w_{cn} \in \mathcal{V} | n=1,\dots , N_c\}$,  with length $N_c$ and vocabulary $\mathcal{V}$. Each utterance has a tag for each of the NLU tasks.
In this work, each utterance has a tag from label sets $\mathcal{L}^{\text{da}}$, $\mathcal{L}^{\text{ic}}$,$\mathcal{L}^{\text{sr}}$ and $\mathcal{L}^{\text{er}}$ indicating dialogue act classification, intent classification, speaker role prediction, and emotion recognition, respectively. This produces a label sequence of the same length $C$ for each task, for instance, $Y^{\text{da}} = \{ y_c^{\text{da}} \in \mathcal{L}^{\text{da}} | c=1,\ldots ,C\}$. Using the maximum a posteriori theory, NLU models seek to output $\hat{Y}^\text{da}$,$\hat{Y}^\text{ic}$,$\hat{Y}^\text{sr}$, and $\hat{Y}^\text{er}$ that maximise the posterior distribution $P(Y^\text{da}|S)$,$P(Y^\text{ic}|S)$,$P(Y^\text{sr}|S)$ and $P(Y^\text{er}|S)$ given $S$, respectively. 

SLU introduces an additional complexity of modeling dialog context from the spoken utterance. Dialog context sequence is formed by $C$ spoken utterances, i.e., $X = \{x_c| c=1,\dots , C\}$. Each spoken utterance $x_c = \{\mathbf{x}_{ct} \in \mathbb{R}^d | t=1,\ldots, T_c\}$ is a sequence of $d$ dimensional speech feature of length $T_c$ frames. Similar to the NLU formulation, SLU systems seek to estimate the label sequence $\hat{Y}^\text{da}$,$\hat{Y}^\text{ic}$,$\hat{Y}^\text{sr}$ and $\hat{Y}^\text{er}$ that maximise the posterior distribution $P(Y^\text{da}|X)$,$P(Y^\text{ic}|X)$,$P(Y^\text{sr}|X)$ and $P(Y^\text{er}|X)$ given $X$, respectively. We can model these posterior distributions as described in the subsections below.

\subsection{Seperate E2E model with dialog context}
\label{Cascaded_Dialog}
Prior work~\cite{ganhotra2021integrating} models each posterior, e.g., $P(Y^\text{da}|X)$, using a sequence of transcripts $S$ by applying the Viterbi approximation:
\begin{align}
    P(Y^\text{da}|X) &= \sum_S P(Y^\text{da}|S,X)P(S|X) \\
    &\approx \max_S P(Y^\text{da}|S,X) P(S|X) \label{eq:cascaded_sum}
\end{align}
Their approach then assumes the conditional independence of $y^\text{da}_c|s_{1:c-1}$ from $x_{1:c-1}$,$y^\text{da}_{1:c-1}$ to simplify the Eq~\ref{eq:cascaded_sum}:
\begin{align}
    P(Y^\text{da}|X) &\approx \max_S \prod_{c=1}^{C} P(y^\text{da}_c | s_{1:c-1}, x_c) P(S|X)  \label{eq:cascaded_CI}
\end{align}
Transcripts are computed using a separate ASR module that seeks to estimate $\hat{S}$ that maximises $P(S|X)$. Using $\hat{S}$, we can modify Eq~\ref{eq:cascaded_CI}:
\begin{align}
    P(Y^\text{da}|X) &\sim \prod_{c=1}^{C} P(y^\text{da}_c | \hat{s}_{1:c-1}, x_c)  \label{eq:cascaded_Shat}
\end{align}
Prior work~\cite{ganhotra2021integrating} models $P(y^\text{da}_c | \hat{s}_{1:c-1}, x_c)$ in Eq.~\ref{eq:cascaded_Shat} by passing ASR transcripts $\hat{s}_{1:c-1}$ to a pretrained language model (LM) like BERT~\cite{BERT} and then concatenating these context embeddings to the acoustic embedding obtained from $x_c$. They focus on building a separate model for each of the SLU tasks, which in the above description is dialogue act classification. Thus, to predict all the SLU tasks, all the separate models that estimate $P(Y^\text{da}|X)$,$P(Y^\text{ic}|X)$,$P(Y^\text{sr}|X)$, and $P(Y^\text{er}|X)$ \textit{independently} need to be executed which can increase latency and computational cost and also does not consider the dependency between SLU tasks.

\section{Proposed Joint E2E model w/ dialog context}
\label{Dialog-integrate}
 In this work, we extend the prior work~\cite{ganhotra2021integrating} on dialog integration discussed in section \ref{Cascaded_Dialog} and propose to jointly model all SLU tasks. We denote a single target containing all the SLU tags as $R=(Y^\text{da},Y^\text{ic},Y^\text{sr},Y^\text{er})$ and modify Eq.~\ref{eq:cascaded_Shat} with $r_c=(y_c^\text{da},y_c^\text{ic},y_c^\text{sr},y_c^\text{er})$ as shown below:
 \begin{align}
    P(R|X) &\sim \prod_{c=1}^{C} P(r_c|\textcolor{blue}{\textbf{$r_{1:c-1}$}},\hat{s}_{1:c-1},   x_c)
 \label{eq:asr_slot}
\end{align}
The SLU tags predicted for the previous spoken utterances i.e. $r_{1:c-1}=(y_{1:c-1}^\text{da},y_{1:c-1}^\text{ic},y_{1:c-1}^\text{sr},y_{1:c-1}^\text{er})$ may be incorporated (Eq.~\ref{eq:asr_slot}) to better model the dialog context unlike in prior work~\cite{ganhotra2021integrating} which assumes conditional independence of $y^\text{da}_c|s_{1:c-1}$ from $y^\text{da}_{1:c-1}$. 
In \Sref{sec:main_res}, we confirm experimentally whether this \textit{previous SLU tag condition} is helpful.
Further, by jointly modeling all the SLU tasks, we expect significantly lower latency and lightweight inference.

\begin{table*}[t]
  \centering
\resizebox {0.91\linewidth} {!} {
\begin{tabular}{l|cccc|cccccc}
\toprule
Model & DA ($\uparrow$) & IC ($\uparrow$) & SR ($\uparrow$) & ER ($\uparrow$) & \multicolumn{2}{c}{RTF ($\downarrow$)} & \multicolumn{2}{c}{Latency (msec.) ($\downarrow$)} & Parameters ($\downarrow$) &  Train time (sec.) ($\downarrow$) \\ 
& & & & & Slowest & Total & Slowest & Total & & \\
\cmidrule(r){6-7} \cmidrule(r){8-9}
 \midrule
Separate E2E model without context* (\cite{sunder2022towards}) & 54.0 & & & & & & & \\ %
Separate E2E model with context* (\cite{sunder2022towards}) & 58.3 & & & & & & &   \\ %
 Separate E2E models without context \Sref{Cascaded_Dialog} & 55.5 & 47.1 & 91.9 & 91.4 & 1.58 & 3.28 & 1153 & 4515 & 457.6M & 4296\\
  \midrule
Joint E2E model without context \Sref{Dialog-integrate} & 55.8 & 47.7 & 91.5 & 90.6 & 1.15 & 1.58 & \hphantom{0}293 & 1153 & 114.4M & 1381 \\
 Joint E2E model with context \Sref{Dialog-integrate} &  58.1 & 86.1 & 95.0 & 90.9 & 1.14 & 1.57 & \hphantom{0}286 & 1133 & 152.7M & 1960\\
\hphantom{000}+ order agnostic training (Eq.~\ref{loss_ctc}) (\textit{proposed}) & \textbf{\underline{58.8}} & \textbf{\underline{86.5}} & \textbf{\underline{95.1}} & \textbf{91.1} & 1.15 & 1.58 & \hphantom{0}289 & 1140 & 152.7M & 2040 \\
\hphantom{000}+ previous SLU tag condition (Eq.~\ref{eq:asr_slot}) (\textit{ablation}) & 58.3 & 86.5 & 94.6 & 90.8 & 1.23 & 1.54 & \hphantom{0}310 & 1080 & 165.4M & 2613\\
\bottomrule
\end{tabular}
}
\vskip -0.1in
\caption{Results presenting the performance of our joint model both with and without using dialog context on dialog act (DA) recognition, intent classification (IC), speaker role (SR) prediction and emotion recognition (ER). * We show DA results of prior work although they have different data preparation setup (sec.~\ref{sec:dataset}). Best joint model results are \textbf{bolded} and further \underline{\textbf{underlined}} if they surpass separate E2E models.} 
\label{tab:main-results}
\vskip -0.15in
\end{table*}
\begin{figure}[t]
  \centering
    \includegraphics[width=0.9\linewidth]{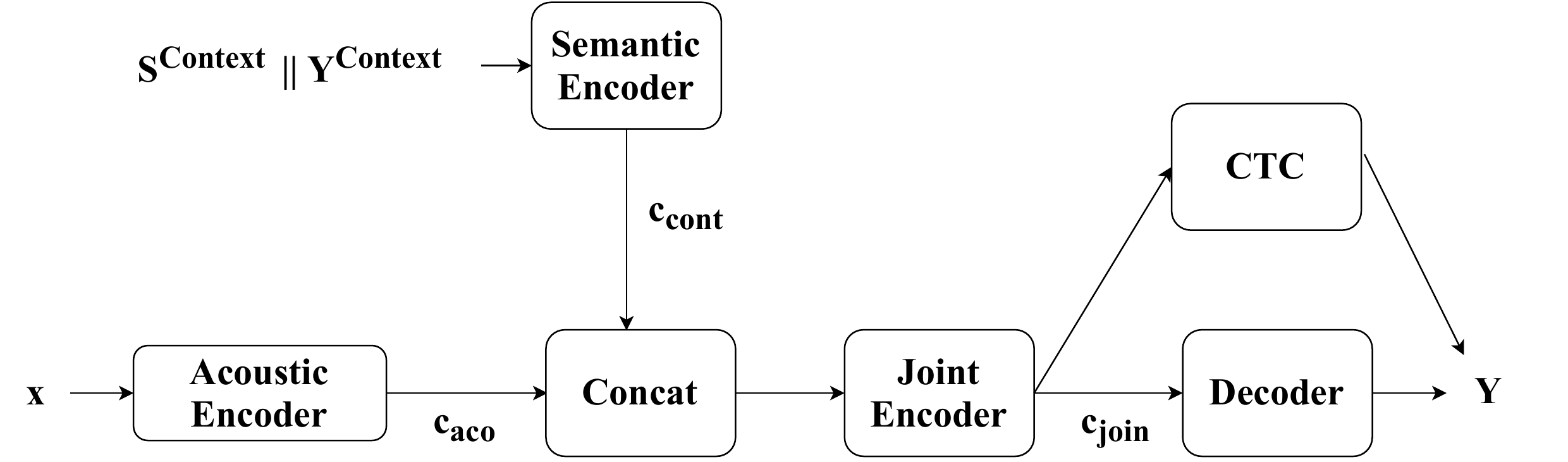}
  \caption{Diagram of our joint E2E model incorporating dialog history for jointly predicting all SLU tasks}
  \label{img:Dialog_SLU}
  \vskip -0.2in
\end{figure}

To realize this formulation, we propose a joint model architecture shown in Fig.~\ref{img:Dialog_SLU}. The input speech signal for each utterance i.e., $x_c$ in Eq.~\ref{eq:asr_slot}, is passed through an acoustic encoder ($\text{Encoder}_{\text{aco}}$) to generate acoustic embeddings $\textbf{c}_{\text{aco}}$.
\begin{align}
&\mathbf{c}_\text{aco} = \operatorname{Encoder}_\text{aco}(x_c)
\label{c_aco}
\end{align}
We concatenate ASR transcripts $\hat{s}_{1:c-1}$ and SLU tags $r_{1:c-1}$ for all previous spoken utterances and pass them through semantic encoder ($\operatorname{Encoder}_\text{sem}$) like a pretrained LM to encode the dialog history:
\begin{align}
&\mathbf{c}_\text{cont} = \operatorname{Encoder}_\text{sem}(\operatorname{concat}(\hat{s}_{1:c-1}, r_{1:c-1}))\label{c_SEM}
\end{align}
The output of the semantic encoder is also passed to a linear layer to ensure that context embeddings $\mathbf{c}_\text{cont}$ have the same hidden dimension as acoustic embeddings $\mathbf{c}_\text{aco}$. The acoustic and context embeddings are concatenated together ($\operatorname{concat}(\mathbf{c}_\text{aco}, \mathbf{c}_\text{cont})$) and attended by a joint encoder $\operatorname{Encoder}_\text{join}$ to produce the joint embedding $\mathbf{c}_\text{join}$:
\begin{align}
&\mathbf{c}_\text{join} = \operatorname{Encoder}_\text{join}(\operatorname{concat}(\mathbf{c}_\text{aco}, \mathbf{c}_\text{cont})) \label{join_output}
\end{align}
The model is trained using joint CTC-attention training~\cite{joint-ctc-att-mtl}, where the CTC objective function is used to train the attention model encoder as an auxiliary task. Because we use an autoregressive decoder to predict tags one by one (Eq.~\ref{l_output}), the likelihood $P(r_c |x_c,\hat{s}_{1:c-1}, r_{1:c-1})$ is dependent on the order of SLU tags in the target sequence $r_c$. This has been referred to as label ambiguity (or permutation) problem in prior work~\cite{weng2015deep,hershey2016deep}. Inspired by prior work on permutation invariant training~\cite{hershey2016deep,pit2,chang2019end}, we use CTC objective function to perform permutation-free training as shown in Eq.~\ref{loss_ctc}, which is referred to as \textbf{order agnostic training} in this work. Let $\mathbf{z}$ be the output sequence variable computed from the joint embedding $\mathbf{c}_\text{join}$, then the optimal permutation order $\hat{\pi}$ is computed as: 
\begin{equation}
    \hat{\pi}=\argmin_{\pi \in \mathcal{P}} \operatorname{Loss}_\text{CTC}(\mathbf{z},r_c^{\pi}),\label{loss_ctc}
\end{equation}
where $\mathcal{P}$ is the set of 4! possible permutations of SLU tags (da,ic,sr,er), $\pi$ is one such permutation and $r_c^{\pi}$ is the reference target with the order of SLU tags indicated by $\pi$. Later, the optimal permutation $\hat{\pi}$ is used for computing the attention decoder loss\footnote{We use the CTC loss to compute optimal order instead of decoder loss for lightweight modeling.}.
\begin{table}[t]
  \centering
\resizebox {0.91\linewidth} {!} {
\begin{tabular}{l|cccc}
\toprule
Model & DA ($\uparrow$) & IC ($\uparrow$) & SR ($\uparrow$) & ER ($\uparrow$)\\ 
 \midrule
Joint E2E Model without context & 54.8 & 47.6 & 91.1 & 90.9\\
\hphantom{000}w/ inter ctc & 55.8 & 47.7 & 91.5 & 90.6  \\
\bottomrule
\end{tabular}
}
\vskip -0.1in
\caption{Results showing the impact of intermediate CTC to advance SLU performance.} 
\label{tab:main-ctc-results}
\vskip -0.2in
\end{table}

For each pair of optimal decoding order and reference target ($\hat{\pi}$,$r_c$), the attention likelihood is calculated as shown below.
\begin{align}
&\mathbf{h}_{l} = \operatorname{Decoder}(\mathbf{c}_\text{JOIN}, (r_c^{\hat{\pi}})_{1:l-1}) \\
&P_\text{Attn}((r_c^{\hat{\pi}})_{l} | x_c, \hat{s}_{1:c-1}, r_{1:c-1}, (r_c^{\hat{\pi}})_{1:l-1}) = \operatorname{SoftmaxOut}(\mathbf{h}_l),
\label{l_output}
\end{align}
where $\operatorname{SoftmaxOut}$ denotes a linear layer followed by the softmax function and $(r_c^{\hat{\pi}})_{l}$ is the $l^{th}$ term in target sequence. The joint likelihood $P_\text{Attn}(r_c^{\hat{\pi}} | x_c, \hat{s}_{1:c-1}, r_{1:c-1})$ can then be computed as a product of the individual likelihood for each of the SLU tags (i.e., from $l$ in [1,4]). It is important to note that this ``order agnostic training'' does not add any new model parameters. During inference, the model automatically picks the tag order, i.e., unlike training, we do not enforce the predicted tag order to have the minimum CTC loss. 
Further, our joint models can also be trained with an auxiliary ASR objective~\cite{ESPnet-SLU,deoras2012joint} by making the model generate both the SLU tags $r_c^{\hat{\pi}}$ and ASR transcript $s_c$.

\section{Experiment Setup}
\label{sec:exp_setup}
\subsection{Datasets}
\label{sec:dataset}
To show the effectiveness of our joint model, we conducted experiments on publicly available HarperValleyBank~\cite{wu2020harpervalleybank} spoken dialog corpus, where dialogs are simulated conversations between bank employees and customers. The corpus consists of 1,446 conversations with 23 hours of audio and 25,730 utterances with human annotated transcripts. The utterances are spoken by 59 unique speakers and have annotations for the dialog act, intent of the conversation, speaker role, and emotional valence.

In this work, we trained a joint model that performs intent classification, dialog act recognition, emotion recognition and speaker attribute prediction. Dialogue act recognition is a multi-label multi-class classification whereas all other tasks are single-label multi-class classification. We followed the setup in prior work~\cite{sunder2022towards} and split the conversations into train, valid and test set\footnote{However, the prior work~\cite{ganhotra2021integrating,sunder2022towards} also uses an off the shelf ASR model to realign the audio with transcripts making their results not directly comparable to results obtained by our setup.}. Similar to ~\cite{ganhotra2021integrating}, we also removed non-lexical tokens such as [noise],[laughter] from the transcript. Our training set contains 1,174 conversations (9.2 hours of audio, 15,424 utterances), valid set contains 73 conversations (0.6 hours of audio, 964 utterances) and our test set contains 199 conversations (1.6 hours of audio, 2903 utterances). We report macro F1 for dialog act prediction and accuracy for intent classification, speaker role prediction and emotion recognition as recommended by prior work~\cite{ganhotra2021integrating,wu2020harpervalleybank}. The latency of our system is reported using two metrics: (1) Real time factor (RTF), which is the average time taken to process an input audio file as a ratio of the duration of input and (2) Endpoint latency which is the elapsed time from the utterance end to getting predictions for SLU tasks. We also show the training time per epoch for all our models.
\begin{table}[t]
  \centering
\resizebox {0.91\linewidth} {!} {
\begin{tabular}{l|cccc}
\toprule
Model & DA ($\uparrow$) & IC ($\uparrow$) & SR ($\uparrow$) & ER ($\uparrow$) \\ 
 \midrule
Joint E2E Model with context &  58.1 & 86.1 & 95.0 & 90.9 \\
Joint E2E Model with oracle context & 58.5  & 87.0 & 95.2 & 91.1\\
\bottomrule
\end{tabular}
}
\vskip -0.1in
\caption{Results showing the robustness of using the ASR transcripts instead of the oracle transcripts.} 
\label{tab:oracle-results}
\vskip -0.2in
\end{table}
\subsection{Architecture details and training}
\label{sec:exp-setup}
Our approach is compared to state-of-the-art SLU task-specific baselines referred to as ``Separate E2E models without context'' (optimizes $P(y^\text{da}_c | x_c)$ instead of $P(y^\text{da}_c | \hat{s}_{1:c-1}, x_c)$ in Eq.~\ref{eq:cascaded_Shat}). We also compared with our proposed model without context, i.e., ``Joint E2E model without context'' which optimizes  $P(r_c |x_c)$ instead of $P(r_c| r_{1:c-1}, x_c, \hat{s}_{1:c-1})$ in Eq.~\ref{eq:asr_slot}. This baseline joint E2E model does not incorporate order agnostic training defined in Eq.~\ref{loss_ctc} and instead is trained using a fixed order $(y^{da}, y^{ic}, y^{sr}, y^{er})$ of SLU tags. 
To understand the efficacy of our proposed modifications to~\cite{ganhotra2021integrating}, we first trained a joint model (referred to as ``Joint E2E model with context'') that does not use SLU tags of previous spoken utterances (i.e. optimize $P(r_c |x_c, \hat{s}_{1:c-1})$ instead of $P(r_c| r_{1:c-1}, x_c, \hat{s}_{1:c-1})$ in Eq.~\ref{eq:asr_slot}) and also does not utilize order agnostic training. In our ablation study, we further trained proposed joint model that incorporates ``order agnostic training'' as defined in Eq.~\ref{loss_ctc} and another joint model that incorporates ``previous SLU tag condition'' i.e. tags predicted for previous spoken utterances ($r_{1:c-1}$ in Eq.~\ref{eq:asr_slot}).

Our models were implemented in pytorch \cite{pytorch} and experiments were conducted through the ESPNet-SLU \cite{ESPnet-SLU,espnet} toolkit. All the models were trained using an auxiliary ASR objective.
Our task-specific baselines consist of 12 layer conformer~\cite{conformer,conformer_another} encoder that inputs features extracted by a strong self-supervised speech model WavLM~\cite{wavlm} and 6 layer transformer~\cite{transformer,transformer_another} decoder. Our ``Joint E2E model without context'' has a similar architecture as task specific baselines.
Prior work~\cite{lee2021intermediate} used an intermediate CTC loss attached to an intermediate layer in the encoder network to regularize CTC training and improve performance. Inspired by this work, we experimented with adding intermediate CTC loss at layers 4, 6 and 8 of our conformer encoder. The ``Joint E2E model without context'' is used to generate dialog context\footnote{The Word Error Rate (WER) of ASR transcripts is 11.7.} $\hat{s}_{1:i-1}$.  We incorporated the usage of Transformers library \cite{wolf-etal-2020-transformers} to get BERT-base-uncased \cite{BERT} as semantic encoder ($\operatorname{encoder}_\text{SEM}$ in Eq.~\ref{c_SEM}).
The conformer architecture with intermediate CTC loss is leveraged as the joint encoder ($\operatorname{encoder}_\text{JOIN}$ in Eq.~\ref{join_output}) in our proposed models. 
Teacher forcing is used for all our models with integrated dialog context i.e., the context was created using ground truth transcripts during training. 
The training and inference of our models were performed using a single NVIDIA Tesla V100-32GB GPU. The inference was computed using 4 parallel jobs and we use the time when the
utterance has been processed by all the jobs (i.e. time taken by slowest job) as well as the sum of time taken by all the jobs to compute latency metrics.
All hyperparameters were
selected based on validation performance.

\section{Results}
\subsection{Main Results}
\label{sec:main_res}
Table~\ref{tab:main-results} shows the results of our joint model both with and without using dialog context. The performance of our ``Separate E2E models without context'' (section~\ref{Cascaded_Dialog}) is similar to the baseline results reported in prior work~\cite{sunder2022towards}, though these results are not directly comparable because of different data preparation setups. Our ``Joint E2E model without context'' performs at par with these task-specific models while significantly reducing latency\footnote{Task-specific baselines are executed one after other in sequential order.}  and the number of trainable model parameters, showcasing the utility of jointly modeling all SLU tasks. We investigated integrating dialog context in the joint model (``Joint E2E model with context'' in Table~\ref{tab:main-results}) using formulation described in section~\ref{Dialog-integrate} and observe a significant improvement in performance, particularly for intent, dialog acts, and speaker role identification, providing evidence that our proposed methodology can effectively encode the context. We further observe a performance gain using ``order agnostic training'', as defined in Eq.~\ref{loss_ctc}, which confirms our hypothesis that the ordering of SLU tags while training can impact model performance. While our order agnostic training framework increases the training time as it requires the computation of CTC loss over all possible permutations of SLU tags (Eq.~\ref{loss_ctc}), this increase in training time is not very significant. 
 We also experimented with ``previous SLU tag condition'' (i.e. attending to $r_{1:c-1}$ in Eq.~\ref{eq:asr_slot}) using only utterance-level annotated tags, i.e., dialog act, emotion, and speaker role, to encode context. 
 This model achieves a performance gain in intent classification and dialog act prediction with comparable results on predicting other SLU tasks. We plan to further investigate the utility of SLU tags from previous spoken utterances in future work.
Our joint E2E model with context also achieved similar performance to ``Separate E2E model with context''~\cite{sunder2022towards} with no knowledge distillation from a text-based system, although the two models have different data preparation setups. 

To better understand our joint model, we also perform an ablation study in Table~\ref{tab:main-ctc-results} and infer that using intermediate CTC loss can stabilize training and improve the performance of our joint model.
\begin{table}[t]
  \centering
\resizebox {\linewidth} {!} {
\begin{tabular}{l|cc|p{5.4cm}}
\toprule
Order & Training (\%) & Inference (\%) & Sample Training Utterance\\ 
 \midrule
 (sr, ic, da, er) & 35.0 & 40.1 & i've ordered your replacement debit card\\
 (da, ic, sr, er) & 27.9 & 36.0 & uh well james my name is jennifer jones and i would like to reset my password please \\
 (ic, sr, da, er) & 16.4 & \hphantom{0}6.8 & thank you for calling have a great day\\
 (ic, da, sr, er) & 14.5 & \hphantom{0}7.5 & you too bye\\
 (sr, da, ic, er) & \hphantom{0}6.2 & \hphantom{0}9.5 & hello this is harper valley national bank my name is patricia how can i help you today\\
\bottomrule
\end{tabular}
}
\vskip -0.1in
\caption{Results showcasing the optimal order of SLU tags found during training using Eq.~\ref{loss_ctc} and predicted order of SLU tags during inference. We report sample training utterance for each tag order.} 
\label{tab:order-results}
\vskip -0.1in
\end{table}

\subsection{Using oracle context}
 To understand the impact of errors in ASR transcripts , we compute results of our joint E2E model with oracle context, i.e., using ground truth transcripts as $\hat{s}_{1:c-1}$ in Eq.~\ref{eq:asr_slot}. Table~\ref{tab:oracle-results} shows only a slight increase in performance compared with the model that uses ASR transcripts to encode context, indicating that it is robust to ASR errors. 

\subsection{Predicted ordering of SLU tags}
\label{sec:optimal_order}
We analyze the optimal order of SLU tags found during training using Eq.~\ref{loss_ctc} in Table~\ref{tab:order-results} and observe that optimal permutation sequence $\hat{\pi}$ are among only 5 of 4! possible permutations of SLU tags. We compare it to the tag order predicted by our joint E2E model with context and order agnostic
training. The predicted order of SLU tags during inference has a similar trend to the optimal order of SLU tags found during training, with  (sr, ic, da, er) and (da, ic, sr, er) being the two most common orders. 
Further the training utterances with optimal order (sr, da, ic, er) are usually spoken at the start of the conversation and are mainly greetings, as shown in Table~\ref{tab:order-results}. We hypothesize that this tag order is optimal for these utterances as it is challenging to detect the intent of the conversation from these greeting utterances, and hence intent is predicted after the model decodes dialogue act and speaker role tags. This gives model the ability to incorporate the dependency on these tags to better extract the intent of current utterance. Our analysis is similar for other tag orders and we infer that we can validate the optimal tag order found for training utterances.   
After a more fine-grained analysis, we observe that test utterances that are similar to training utterances with optimal tag order $\pi$ are also predicted by the joint model in the same tag order $\pi$ during inference. This finding provides initial evidence to the hypothesis that the joint model learns to automatically predict in the optimal tag order during inference using order agnostic training. Based on this interesting insight and performance gains observed in Table~\ref{tab:main-results}, we recommend future studies on jointly modeling different SLU tasks to incorporate order agnostic training in their framework.

\section{Conclusion}
We propose a novel model architecture that can jointly model intent classification, dialogue act prediction, speaker role identification and emotion recognition with full integration of dialog history in spoken conversations. Our results show that the joint model achieves comparable performance to task-specific models with the additional benefits of low latency and lightweight inference. Our joint model can also successfully capture dialog context to improve the prediction performance of all SLU tasks significantly. We experimentally confirm that order agnostic training can further enhance performance. In future work, we plan to explore E2E integration of dialog context as well as knowledge distillation from a text-based system.
\section{Acknowledgement}
This work used the Extreme Science and Engineering Discovery Environment (XSEDE) \cite{xsede}, which is supported by NSF grant number ACI-1548562. Specifically, it used the Bridges system \cite{nystrom2015bridges}, which is supported by NSF award number ACI-1445606, at the Pittsburgh Supercomputing Center (PSC).

\section{References}
{
\printbibliography
}
\end{document}